\documentclass[letterpaper, 10 pt, conference]{ieeeconf}  

\IEEEoverridecommandlockouts                              
\usepackage{amsmath} 
\usepackage{amssymb}  

\usepackage{siunitx}
\usepackage{hyperref}
\usepackage[font=small]{caption}
\usepackage{graphicx}
\usepackage{comment}
\usepackage{xcolor}

\usepackage{numprint}
\npthousandsep{,}\npthousandthpartsep{}\npdecimalsign{.}

\newcommand{\new}[1]{\textcolor{black}{#1}}
\newcommand{\del}[1]{\iffalse {#1} \fi}

\newcommand{\n}[1]{\textcolor{black}{#1}}
\newcommand{\w}[1]{\iffalse {#1} \fi}

\newcommand{\rn}[1]{\textcolor{black}{#1}}
\newcommand{\rw}[1]{\iffalse {#1} \fi}

\title{\LARGE \bf
Computationally Efficient Reinforcement Learning:\\Targeted Exploration leveraging Simple Rules
}

\author{Loris Di Natale, Bratislav Svetozarevic, Philipp Heer, and Colin N. Jones
\thanks{This research was supported by the Swiss National Science Foundation under NCCR Automation, grant agreement 51NF40\_180545, and in part by the Swiss Data Science Center, grant agreement C20-13.}%
\thanks{L. Di Natale, B. Svetozarevic, and P. Heer are with the Urban Energy Systems Lab, Empa, D\"{u}bendorf, Switzerland.
        {\tt\small \{loris.dinatale, bratislav.svetozarevic, philipp.heer\}@empa.ch}.}%
\thanks{L. Di Natale and C. N. Jones are with the Laboratoire d'Automatique, EPFL, Lausanne, Switzerland.
        {\tt\small colin.jones@epfl.ch}.}%
}

\begin{document}

\maketitle
\thispagestyle{empty}
\pagestyle{empty}


\begin{abstract}

\rn{Model-free} Reinforcement Learning (RL) generally suffers from poor sample complexity, mostly due to the need to exhaustively explore the state\n{-action} space to find \n{well-performing}\w{good} policies. On the other hand, we postulate that expert knowledge of the system \del{to control }often allows us to design simple rules we expect good policies to follow at all times. In this work, we hence propose a simple yet effective modification of continuous actor-critic \del{RL }frameworks to incorporate such \new{rules}\del{prior knowledge in the learned policies} and \w{constrain RL agents}\del{them}\w{ to}\n{avoid} regions of the state\new{-action} space that are \w{deemed interesting}\n{known to be suboptimal}, thereby significantly accelerating the\w{ir} convergence \n{of RL agents}. Concretely, we saturate the actions chosen by the agent if they do not comply with our intuition and, critically, modify the gradient update step of the policy to ensure the learning process is not affected \n{by} the saturation step. On a room temperature control \del{simulation }case study, \new{it}\del{these modifications} allow\new{s} agents to converge to well-performing policies up to \new{$6-7\times$}\del{one order of magnitude} faster than classical \del{RL }agents \new{without computational overhead and} while retaining good final performance.

\end{abstract}


\section{Introduction}

\new{Despite its success}\del{While being successful} in many applications \cite{coronato2020reinforcement, 
zhang2019deep}, \rn{model-free} Reinforcement Learning (RL), and in particular \new{deep RL}\del{its deep counterpart} (DRL), usually suffer from \textit{data inefficiency}, i.e., they \del{typically }require a significant number of interactions with the environment to converge \cite{duan2016rl, schwarzer2021pretraining}. This \new{stems from the necessity to explore the state-action space to find optimal policies and}\del{ not only} leads to significant computational costs\del{ but also}. \new{It also} limits the deployment of DRL methods on physical systems without pretraining in simulation\del{. Indeed,}\new{:}\del{in the case of building temperature control, for example,} learning a \new{building temperature control} policy from scratch can \new{for example} take years of data \cite{wang2020reinforcement, di2022near}. 

\new{To speed up the training of DRL agents, researchers have for example}\del{ designed custom exploration strategies \cite{lipton2018bbq} or} \new{investigated how to leverage expert demonstrations \cite{nair2018overcoming, hester2018deep}, but this requires access to an expert policy which is not always available in practice.} \del{On the other hand}\new{Instead}, we postulate that prior knowledge of physical systems often allows us to design simple rules that \del{RL }agents should follow \textit{a priori}, such as \textit{``Do not heat the room if it is already \SI{26}{\celsius}''}\new{; we indeed know this action will \textit{always} be suboptimal in that state, there is no need for\del{ DRL} agents to explore its consequences}. \del{Interestingly, incorporating such \del{expert }knowledge in control policies is a promising step towards more efficient physics-informed RL algorithms~\cite{du2022demonstration}.} 

In this paper, we hence propose modifications of actor-critic algorithms to encode simple \w{\new{physical} }rules in RL agents, introducing \textit{\del{artificial }\new{state-dependent} constraints} on the \new{agents'} actions\del{ taken by the agents} to restrict exploration to interesting regions of the state\new{-action} space\new{. In other words, the key idea is to avoid visiting state-action pairs that are known to be suboptimal by the expert to accelerate the convergence towards meaningful solutions and thus increase the efficiency of (D)RL. Note that while state-dependent bounds were concurrently introduced in~\cite{de2023enforcing}, they are enforced \textit{a posteriori} in the environment instead of directly modifying the agent's behavior and do not necessarily improve convergence. In another line of work, prior knowledge successfully accelerated learning in \cite{zhang2020kogun}, but relying on fuzzy rather than direct rule integration. }

\subsection{\new{Constraining RL agents}}

\new{To bound the decisions taken by RL agents, one typically defines some \textit{constrained set} of actions and either project the actions of the agents on this set at each time step or switch to a fallback controller when needed~\cite{wangtoward, mao2019towards, bautista2022autonomous, goh2022assessment}. The main challenge with these operations is that they are usually not differentiable and hence cannot be learned by RL agents, with the notable exceptions of \cite{chen2021enforcing, gros2020safe, dalal2018safe}, who leveraged differentiable optimization layers \cite{chen2021enforcing}, modified the policy updates to account for projections \cite{gros2020safe}, or 
derived a closed-form solution of the projection step~\cite{dalal2018safe}.}\del{ The linear assumption was then lifted in \cite{wangtoward}, but only to correct the actions of the agents online, as it cannot be used in training since no closed-form solution exists anymore.} \new{However, these methods either entail additional computational burden~\cite{chen2021enforcing, gros2020safe} 
or rely on a learned linearization of the constraints~\cite{dalal2018safe}.} 
\new{Alternatively, one could also \n{apply}\w{borrow} tools from the safe RL literature~\cite{osinenko2022reinforcement, gu2022review, garcia2015comprehensive}, typically relying on constrained policy optimization~\cite{simao2021alwayssafe, achiam2017constrained}. However, this would again introduce both engineering and computational overhead.} 

\new{The complexity of the methods discussed above often \n{stems}\w{comes} from the fact that they are designed to impose \textit{state constraints} on DRL agents, which is a more challenging problem in general since it leads to complex action bounds for the agent at each step. Here, however, we argue that prior knowledge can straightforwardly be used to accelerate the training of DRL agents through simple \n{state-dependent} \textit{box constraints on their actions}, which allows us to leverage \w{less principled but also }less computationally expensive tools.}

\del{While many works rely on the definition of (potentially complex) \textit{constrained} set of actions, projecting the decisions of the agents on this set at each step to saturate them, this operation is often either not differentiable \cite{bautista2022autonomous} or computationally intensive~\cite{chen2021enforcing, gros2020safe}. To alleviate the issue of non-differentiability without additional computational load, the main focus of this paper, one can instead let agents learn when to switch to the fallback controller~\cite{xie2018learning, hsu2022sim}, but the satisfaction of the constraints cannot be guaranteed any more. Closer to our work}
\new{To alleviate the issue of non-differentiability without increasing either the engineering or the computational burden}, Reward Shaping (RS) \new{heuristics} might be used in various forms to penalize agents when constraints are violated, let them know when a fallback controller was used or they were saturated\del{ instead of following their choice}, or \new{introduce}\del{include} prior knowledge \new{about the task to solve}\del{to accelerate training} \del{even when the saturation step is not differentiable }\cite{goh2022assessment, alshiekh2018safe, hu2020learning}. While such methods \new{might accelerate the learning process to some extent}, they are however indirect, i.e., they \new{only} influence the learned policies through the reward function that the agent will learn to optimize over time. \new{Moreover, shaping the reward function simultaneously impacts the learning process of both the actor \textit{and} the critic.}


\subsection{\new{Contribution}}

\del{In this work, we propose to saturate DRL agents' actions when they are deemed uninteresting according to prior expert knowledge, which is encoded in simple rules. We then modify the actor update step}\new{In this work, we propose to clip DRL agents' actions according to simple expert-designed state-dependent bounds and then modify the gradient update step of the actor} to let agents learn from their mistakes and \n{accelerate their convergence to}\w{steer their decisions towards} expected actions. \new{To explain the effectiveness of the gradient modification, we provide an intuitive analytical analysis of its impact on the learning of DRL agents. Importantly, contrary to RS, our modifications only \n{affect}\w{impact} the actor, allowing the critic to learn the true Q-values.}

\new{Remarkably, our method bypasses the need for complex projection steps and does not require access to a fallback controller or an expert policy. 
Moreover and c}ritically, the proposed modifications do not impact the \new{computational} complexity of the \del{learning }algorithm, are straightforward to design and implement, \n{and} can be coupled with \new{any}\del{most existing} actor-critic algorithms.\w{, and enforce the wanted behaviors on DRL agents \textit{by design}. \w{This is in stark contrast with \cite{zhang2020kogun}, where the expert knowledge is distilled and modified by the control policy.}}
\n{Note that, in contrast with~\cite{zhang2020kogun}, where the expert knowledge is potentially overridden by the policy, our method enforces the wanted behaviors on agents \textit{at all times}.}

\w{Their effectiveness}\n{The effectiveness of the proposed Efficient Agents (EAs)} is demonstrated in simulation on a room temperature control case study, where \w{the proposed Efficient Agents (EAs)}\n{they} converge up to $6$ -- $7$ times faster than classical ones \new{and $2$ -- $3$ times faster than RS-based \n{agents}\w{learning algorithms}} while retaining good final performance. \new{This \rn{hints at}\rw{shows} how the proposed modifications \rn{can} provide a simple yet effective and computationally inexpensive mean to leverage expert knowledge to accelerate DRL algorithms.}





\section{Preliminaries}

\subsection{Reinforcement Learning}

At each time step $t$, given an observation $s_t$ of the state of the environment, an RL agent chooses an action $a_t$. The environment then transitions to $s_{t+1}$ according to the transition probabilities $P(s_t, a_t)$ and sends the new state and the reward signal $r(s_t, a_t)$ to the agent. The objective of any \new{deterministic} RL algorithm\footnote{While the presented analyses deal with deterministic actor-critic agents for clarity, the results can easily be extended to the stochastic case.} is to find a policy $\pi(s_t)$ that maximizes the expected discounted cumulative returns:
\begin{align}
    J(\pi) = \mathbb{E}_{a_t\sim\pi(s_t), s_{t+1}\sim P(s_t, a_t)}\left[\sum_{t=0}^{\infty}{\gamma^tr(s_t, a_t)}\right], \label{equ:obj}
\end{align}
where $\gamma$ is the discount factor trading off near- and long-term rewards, and the initial state $s_0\sim\rho$ is sampled from the corresponding initial distribution. With a slight abuse of notation on the expectation for clarity, we can define the Q-function of any state-action pair $(s,a)$: 
\begin{align}
    Q^\pi(s,a) = \mathbb{E}_{\pi}\left[\sum_{t=0}^{\infty}{\gamma^tr(s_t, a_t)|s_0=s, a_0=a}\right], 
\end{align} 
which captures the expected returns when action $a$ is chosen in state $s$ and the policy $\pi$ is followed thereafter.

In our simulations, we let agents explore the environment with the $\epsilon$-greedy exploration strategy, which means we apply the following action to the environment:
\begin{equation}
    a(s) = \text{clip}(\pi(s)+\epsilon, a^{\textit{low}}, a^{\textit{up}}), \qquad \epsilon \sim \mathcal{N}(0,\sigma), \label{equ:action}
\end{equation}
where the noisy actions are clipped \n{elementwise} between $a^{\textit{low}}$ and $a^{\textit{up}}$, the predefined action bounds \new{from the environment}, and $\epsilon$ is the Gaussian exploration noise with a standard deviation of~$\sigma$. All the transition tuples $(s,a,r,s')$ observed by the agent are stored in a replay buffer.

\subsection{Actor-critic algorithms}

In practice, policies and Q-functions are often parametrized with Neural Networks (NNs) as $\pi_\theta$ and $Q_\phi$, respectively, leading to DRL, and numerous algorithms have been developed to maximize~\eqref{equ:obj} \cite{openai}. In this work, we are interested in deterministic actor-critic methods stemming from \cite{lillicrap2015continuous}, where both the actor $\pi_\theta$ (also referred to as the policy) and the critic $Q_\phi$ are optimized in parallel \del{with}\new{leveraging} gradient descent. While different flavors exist, most \del{actor-critic }algorithms compute the gradient of the critic using \rw{a variant of }the Temporal Difference (TD) loss\rn{~\cite{lillicrap2015continuous}:}\rw{and use the critic to estimate the expected returns and derive \new{the }actor gradient\del{s} as~\cite{lillicrap2015continuous}:} 
\begin{align}
    \hat\nabla_{\phi} Q_{\phi} &= \nabla_{\phi} \left[\frac{1}{|B|}\sum_{b\rw{:}=(s,a,r,s')\in B}{\left(Q_{\phi}(s,a) - y(b)\right)^2}\right], \label{equ:criticclassical}
    \rw{\hat\nabla_{\theta} \pi_{\theta} &= - \nabla_{\theta} \left[\frac{1}{|B|}\sum_{s \in B}{Q_{\phi}(s,\pi_\theta(s))}\right],}
\end{align}
\new{with $y(b) = \left(r + \gamma\max_{a'}  Q_{\phi}(s',a')\right)$ and }where a batch $B$ of past transitions is sampled from the replay buffer and used to estimate expectations\rn{. }\rw{, \eqref{equ:criticclassical} is the TD loss, and \eqref{equ:actorclassical} is verified by}\rn{Leveraging} the policy gradient theorem~\cite{silver2014deterministic}\rn{, one can similarly use the critic to estimate the actor gradient as:}
\begin{align}
    \rn{\hat\nabla_{\theta} \pi_{\theta}} &= - \rn{\nabla_{\theta} \left[\frac{1}{|B|}\sum_{s \in B}{Q_{\phi}(s,\pi_\theta(s))}\right].} \label{equ:actorclassical}
\end{align}
Note that these gradients are easily computed using automatic differentiation when the actor and the critic are parametrized with NNs. \del{, see for example in \cite{paszke2017automatic}.}

In this paper, we rely on the Twin Delayed Deep Deterministic (TD3) policy gradient algorithm, which introduces a few modifications to limit the well-known overestimation bias of Q-functions plaguing vanilla actor-critic algorithms~\cite{fujimoto2018addressing}. Remarkably, however, these adjustments do not impact the actor gradient in \eqref{equ:actorclassical}, \new{allowing}\del{which allows} us to seamlessly integrate the \new{proposed} modifications detailed in Section~\ref{sec:methods}. \del{ into TD3.}


\section{Methods}
\label{sec:methods}

\w{In RL applications, prior \del{or }expert knowledge often allows us to formulate simple rules that any well-performing policy should follow, such as \textit{``\new{Do not turn left if there is a wall on your left}\del{Immediately heat the room if the temperature is \SI{16}{\celsius}}''}. This section details how to encode such simple rules in actor-critic algorithms to limit the exploration of \new{known suboptimal state-action pairs}\del{uninteresting states}\n{;}\w{,} first\del{ly by} saturating actions taken by the control policy \new{accordingly} and then \del{by }modifying the gradient \del{update} of the actor \del{accordingly} to let agents learn from their mistakes.}

\subsection{\new{State-dependent a}\del{A}ction saturation}

In many cases, prior knowledge allows us to design state-dependent upper and lower bounds $a^{\textit{max}}(s)$ and  $a^{\textit{min}}(s)$, respectively, on the actions we expect \n{well-performing}\w{good} \new{control} policies to take in \new{a given state $s$, \n{with}\w{where}}\del{each state, with}: 
\begin{align}
   \quad a^{\textit{low}} \leq a^{\textit{min}}(s) \leq a^{\textit{max}}(s) \leq a^{\textit{up}}. \label{equ:bounds}
\end{align}
\n{To limit the exploration of known suboptimal state-action pairs, w}\w{W}e can then \new{modify}\del{clip the (noisy) action chosen by the policy in} \eqref{equ:action} \new{accordingly to\w{ restrict agents}}\del{between the provided bounds to restrict exploration to interesting parts of the state space}: 
\begin{align}
    a(s) &= \text{clip}(\pi_\theta(s)+\epsilon, a^{\textit{min}}(s), a^{\textit{max}}(s)). \label{equ:newaction} 
\end{align}
Note that these bounds stemming from prior knowledge are also enforced at test time when $\epsilon=0$ \n{to ensure an agent would \textit{never} turn left if whenever there is a wall in that direction, for example, neither during the training nor the deployment phase}.

\subsection{\del{Modified actor gradients}\new{Actor gradient modification}}

\new{The major problem with \n{the clipping operation in}~\eqref{equ:newaction} is \rw{that is not differentiable}\rn{its non-differentiability}. 
Worse yet, its subdifferentials go to zero whenever agents are saturated (see~\eqref{equ:gradient} in Section~\ref{sec:theory}), making any backward flow of information on the overriding process impossible.} \del{One major problem when saturating the actions of the policy as suggested above is that this operation is not differentiable and the subdifferentials are zero if actions are saturated (see \eqref{equ:gradient} below). Hence, we cannot backpropagate through it to let the policy learn from its mistakes.} As a countermeasure\new{, to let agents learn from their mistakes}, we also modify the actor gradient \eqref{equ:actorclassical} \new{to}\del{as follows}:
\begin{align}
    \hat\nabla^{\textit{EA}}_{\theta} \pi_{\theta} = - \nabla_{\theta} \Bigg(&\frac{1}{|B|}\sum_{(s,a) \in B}\bigg[{Q_{\phi}(s,\pi_\theta(s))} \nonumber \\
    & - \frac{\lambda}{2} \left(\pi_\theta(s) - a(s)\right)^2\bigg]\Bigg), \label{equ:actormodified}
\end{align}
where $\lambda$ is a \del{weighting}\new{hyper}parameter. The \new{last}\del{additional} term in \eqref{equ:actormodified} penalizes actions chosen by the policy $\pi_\theta(s)$ if they deviate from the constrained action $a(s)$ that was applied to the environment, thus steering the agent's decisions towards expected actions.\footnote{\new{In another line of work, this penalty was also used in~\cite{chen2021enforcing} to improve the robustness of differentiable layer-based RL for state-constrained problem\n{s}.}}

\new{Alternatively, we note here that one could instead modify the reward function \w{(RS) }to include this penalty \n{(RS)} and then maximize \eqref{equ:obj}. Remarkably, however, the latter also impacts the learning process of the critic in~\eqref{equ:criticclassical} when applied to actor-critic frameworks, contrary to our method. We will show empirical benefits of the proposed modification~\eqref{equ:actormodified} over RS-based penalties in terms of convergence speed in Section~\ref{sec:results}.}

\del{Note that this modification was also successfully applied in \cite{chen2021enforcing}. In this work, however, we postulate that the simple gradient modification \eqref{equ:actormodified} can be enough to accelerate the training of DRL agents on physical systems, in parallel to the differentiable projection on a safe set of actions, and shown to improve the convergence speed of the agents. In this work, however, we postulate that the simple gradient modification \eqref{equ:actormodified} can be enough to accelerate the training of DRL agents on physical systems, alleviating the computational burden of differentiable projection layers.}

\subsection{Implications of the modified gradients}
\label{sec:theory}

Let $C(s) = \left\{a\in\mathbb{R}: a^{\textit{min}}(s) \leq a \leq a^{\textit{max}}(s)\right\}$ \new{for}\del{in} any given state $s$.\footnote{\new{Without loss of generality, we assume that $a\in\mathbb{R}$ in this section for clarity. This assumption can easily be lifted for multi-dimensional problems.}} \del{Relying on the definition of the clipping operator \eqref{equ:newaction} and g}\new{G}rouping all the parameters $\theta$ in a vector \n{and recalling the definition of the action $a(s)$ applied to the environment in state $s$ from~\eqref{equ:newaction}}, we can define \w{the}\n{its} subgradient $\nabla_\theta a(s)$ \n{as}\w{of the action $a(s)$ applied to the environment in state $s$ as follows based on~\eqref{equ:newaction}}: 
\begin{align}
    a(s) &= \begin{cases}
    a^{min}(s), & \text{if } \pi(s) < a^{min}(s),\\
    \pi_\theta(s) + \epsilon, & \text{if } \pi_\theta(s)\in C(s),\\
    a^{max}(s), & \text{if } \pi(s) > a^{max}(s).
    \end{cases} \nonumber \\
    \implies \nabla_\theta a(s) &= \begin{cases}
    \nabla_\theta\pi_\theta(s),  & \text{if }
    \pi_\theta(s)\in C(s), \\
    0, & \text{else,}
    \end{cases} \label{equ:gradient}
\end{align}
\del{Introducing the Jacobian $\nabla_{\theta}\pi_\theta(s)$, w}\new{where $\nabla_{\theta}\pi_\theta(s)$ is the \n{actor} gradient. W}e can \new{then} rewrite \n{the gradient of EAs~\eqref{equ:actormodified}} as:
\begin{align*}
    \hat\nabla^{\textit{EA}}_{\theta} \pi_{\theta} &= - \frac{1}{|B|}\sum_{(s,a) \in B} \bigg[ {\nabla_{\theta} Q_{\phi}(s,\pi_\theta(s))} \nonumber \\
    & \qquad \quad - \nabla_{\theta}\left(\frac{\lambda}{2} \left(\pi_\theta(s) - a(s)\right)^2\right)\bigg] \\
    &= - \frac{1}{|B|}\sum_{(s,a) \in B} \bigg[ \nabla_{\theta} Q_{\phi}(s,\pi_\theta(s)) \\
    & \qquad \quad - \left( \lambda \left(\nabla_{\theta}\pi_\theta(s) -  \nabla_{\theta}a(s)\right)^\top\left(e_\theta(s)\right)\right)\bigg],
\end{align*}
where we introduce the error \n{term} $e_\theta(s) = \pi_\theta(s) - a(s)$. We hence get the following modified actor gradient\w{ \new{for EAs}}, where we omit $(s,a) \in B$ for clarity:
\begin{align*}
    \hat\nabla^{\textit{EA}}_{\theta} \pi_{\theta} &= \begin{cases} - \frac{1}{|B|}\sum_{B}\Big[ \nabla_{\theta} Q_{\phi}(s,\pi_\theta(s))\Big], \ \text{if } \pi_\theta(s)\in C(s), \\
    - \frac{1}{|B|}\sum_{B} \Big[ \nabla_{\theta} Q_{\phi}(s,\pi_\theta(s)) \\
    \qquad \quad \  - \lambda \nabla_{\theta}\pi_\theta(s)^\top e_\theta(s) \Big], \ \text{else.}\end{cases} \label{equ:impactanalyzed} 
\end{align*}
Remarkably, the additional penalty term \new{in~\eqref{equ:actormodified}} \n{hence} allows us to solve the issue of the subdifferentials of the clipping operator being zero when actions are saturated, modifying the gradients \del{exactly}\textit{\new{only} when the constraints are not met}. Indeed, as long as the action chosen by the agent respects the constraints provided by the expert, the classical gradient~\eqref{equ:actorclassical} is used. On the other hand, as soon as the constraints are not met, the gradient is modified in the direction $e_\theta(s)$ to accelerate the convergence of $\pi_\theta(s)$ to $C\new{(s)}$ despite the subdifferential of the clipped action being zero\new{, confirming the graphical intuition from~\cite[Fig. 2]{chen2021enforcing}.} This allows \n{EAs}\w{agents} to learn from their mistakes and \n{---} \new{we hypothesize \n{---} helps them} rapidly converge to \new{meaningful}\del{interesting} policies.


\section{Room temperature control case study}

To assess the effectiveness of the proposed method, we apply it to a \new{temperature}\del{building} control case study, where the objective is to minimize the energy consumption of a room while maintaining the comfort of the occupants, represented by predefined temperature bounds that should not be exceeded.

\subsection{Reinforcement Learning framework}

The continuous action space of the agents corresponds to how much heating \new{or}\del{ power, respectively} cooling power, should be applied at each time step, normalized between $a^{\textit{low}}=-1$ and $a^{\textit{up}}=1$. During the heating season, $a^{\textit{low}}$ corresponds to the heating being turned off\del{,} and $a^{\textit{up}}$ to full heating, and the contrary in the cooling case. Physically Consistent Neural Networks (PCNNs) \cite{di2022physically} are used to simulate \del{the behavior of }one bedroom in the NEST building~\cite{nest} \new{and $s_t$ gathers time, weather, temperature, and comfort bound information (see~\cite{di2022near} for details)}. The reward function is defined as the negative weighted sum of energy consumption $E_t$ and comfort violations, i.e. how far from the designed bounds the temperature inside the room is: 
\begin{align}
    r(s_t,a_t) &= - \max{\{L_t-T_t, T_t-U_t, 0\}} - \alpha E_t, \\
    E_t &= \begin{cases} \frac{a_t + 1}{2}\ E^{\textit{max}}_{\textit{heat}}, & \text{in the heating season},\\ 
    \frac{1 - a_t}{2}\ E^{\textit{max}}_{\textit{cool}}, & \text{in the cooling season}. \end{cases} \nonumber
\end{align}
where $L_t$ and $U_t$ represent the lower and upper comfort bounds on the temperature $T_t$ at time $t$, respectively, $\alpha$ is a weighting factor, and $E^{\textit{max}}_{\textit{heat}}$ and $E^{\textit{max}}_{\textit{cool}}$ stand for the maximal heating and cooling power\new{, respectively}. \del{More details on the experimental setup can be found in \cite{di2022near}.}

\subsection{Design of the saturation rules}
\label{sec:f}

\begin{figure}[]
  \begin{center}
    \centerline{\includegraphics[height=0.3\columnwidth]{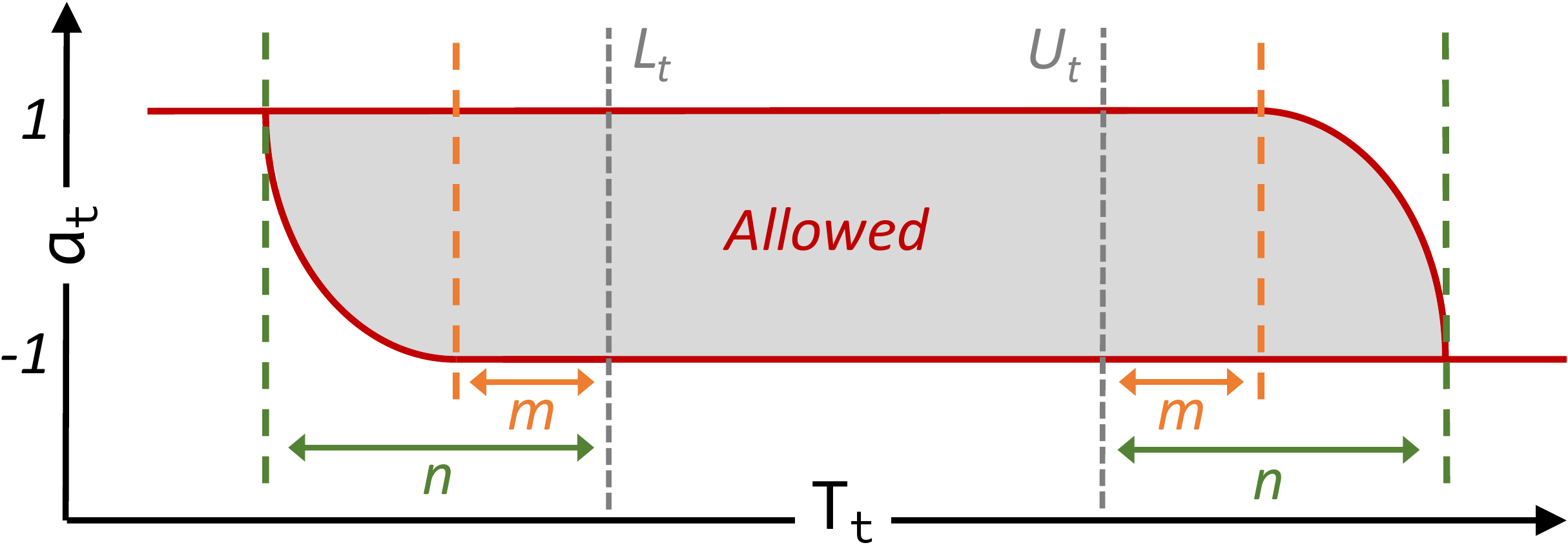}}
  \end{center}
  \vspace{-1.5em}
  \caption{\new{Representation of the action bounds used in this work.}}
    \label{fig:bounds}
\vspace{-1.5em}
\end{figure}

In the context of room temperature control, we intuitively know that an optimal policy should \n{gradually}\w{slowly} stop heating when the temperature reaches the upper comfort bound and \n{gradually}\w{slowly} start heating as soon as the lower bound is not met (and \new{vice versa}\del{the contrary} for cooling)\del{, typically to avoid criticism from the occupants}. To encode these simple rules, we design \new{state-dependent action bounds}\del{$f$} as follows:
\begin{align}
    a^{\textit{min}}(s_t) &= \text{clip}\left(\frac{(L_t - m) - T_t}{n - m},\ 0,\ 1\right)^2 * 2 - 1 \\
    a^{\textit{max}}(s_t) &= 1 - 2 * \text{clip}\left(\frac{T_t - (U_t + m)}{n - m},\ 0,\ 1\right)^2,
\end{align}
with $n \geq m \geq 0$ representing design parameters to leave more or less freedom to the agents. In words, we start constraining the action of the agents as soon as the temperature deviates from the bounds for more than $m$ degrees and then quadratically increase the constraint until $n$ degrees have been reached, where the agent is forced to use the maximum or minimum power\new{, as pictured in Fig.~\ref{fig:bounds}}. \del{As can be seen, $a^{\textit{min}}(s_t)>-1$ only when the temperature is below the lower comfort bound, and $a^{\textit{max}}(s_t)<1$ only when it exceeds the upper one.} 

\section{Results}
\label{sec:results}

\new{To investigate the influence of $m$ and $n$, which measure how much prior knowledge is transmitted to DRL agents, w}\del{W}e train\del{ed} \w{several}\n{different} EAs\del{ with different parameters} (EA\;$m$\;/\;$n$)\del{ to minimize the energy consumption while maintaining the comfort of the occupants}. For comparison purposes, we also train\del{ed} agents with the classical actor gradient \eqref{equ:actorclassical}, introducing the additional squared penalty $\frac{\lambda}{2} \left(\pi_\theta(s) - a(s)\right)^2$ in the reward function instead as another computationally inexpensive means to incorporate prior knowledge in \del{DRL }agents (RS\;$m$\;/\;$n$). \new{Finally, we also analyze two classical DRL agents with different random seeds (Classical $1$ and $2$).\footnote{The code and data\del{ used to generate these results} are available on \url{https://gitlab.nccr-automation.ch/loris.dinatale/efficient-drl}.}}

\subsection{\new{Final performance}}
\label{sec:perf}

All the agents were trained on up to three-day-long episodes randomly sampled from three years of data. They were evaluated after each $96$ steps of \SI{15}{\minute}, i.e. one day's worth of data, hereafter also referred to as an \textit{epoch}, on a testing set of $50$ unseen sequences of three days\del{ to monitor their progress during the first $500$ epochs}. \new{They all use the same hyperparameters as in~\cite{di2022near}. We manually set $\lambda=100$ for EAs to ensure the constraints are enforced as fast as possible and $\lambda=10$ for RSs since larger penalties led to instability. While we empirically observed more robust performance of EAs with respect to $\lambda$ compared to RSs, a complete sensitivity analysis is left for future work.}

\begin{figure}[]
  \begin{center}
    \centerline{\includegraphics[width=\columnwidth]{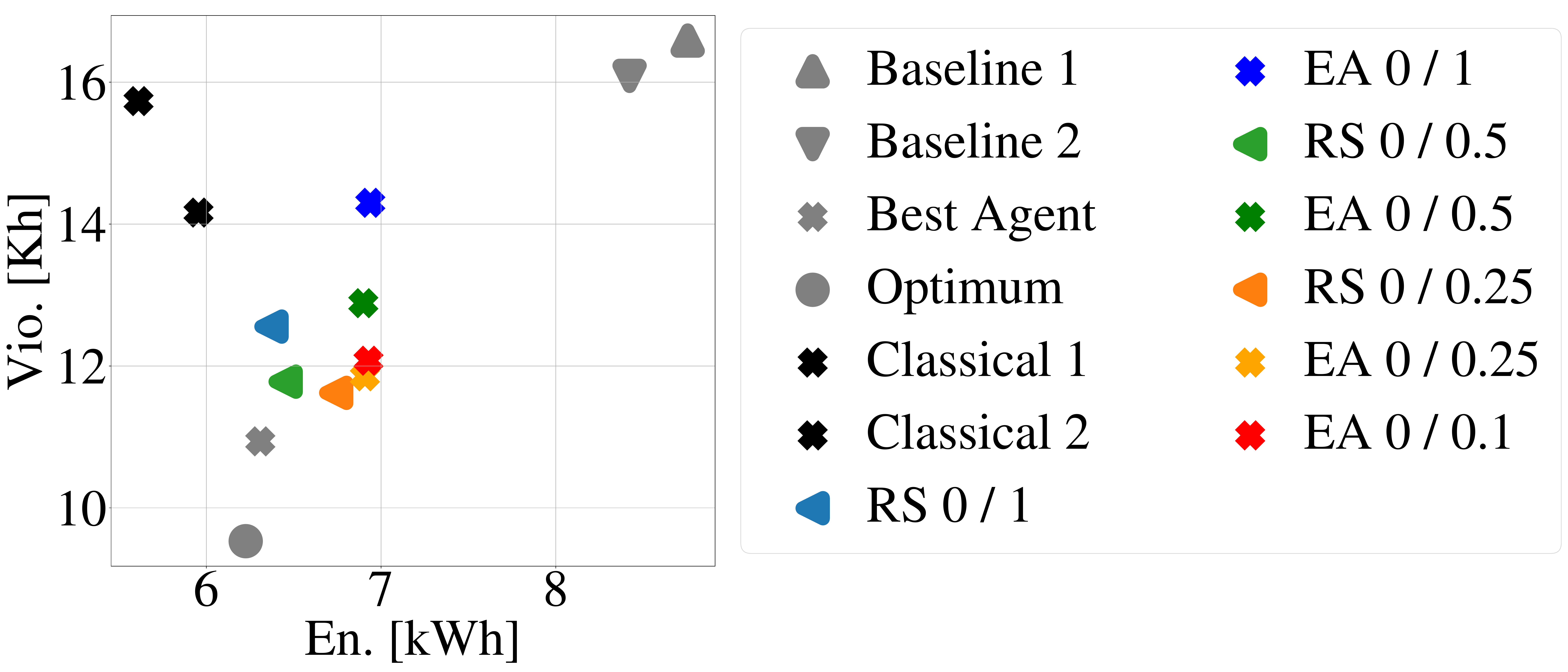}}
  \end{center}
  \vspace{-1.5em}
  \caption{\del{Best trade-off between e}\new{Average e}nergy consumption (En.) and comfort violations (Vio.) \new{over the test set corresponding to each agent in Table~\ref{tab:final}.}\del{found by different agents on the testing set over the first $500$ epochs.} \del{For comparison purposes, t}\new{T}he performance of two industrial baselines, of an agent trained for \numprint{125000} epochs (Best Agent), and the optimal performance achievable \new{(Optimum)}, all computed as in \cite{di2022near}, are\del{ also} reported in gray.}
  \label{fig:results_app}
    \vspace{-1.5em}
\end{figure}

\new{The best reward obtained by all the trained agents over the first $500$ epochs can be found in Table~\ref{tab:final}, and the corresponding trade-off between energy consumption and comfort violations is plotted in Fig.~\ref{fig:results_app}. These results illustrate how tighter \n{parameters $m$ and $n$}\w{constraints $m$\;/\;$n$}, i.e., higher levels of prior knowledge, allow EAs and RSs to converge to better solutions in this limited training regime. In particular, it allows EAs to reduce the amount of comfort violations without significantly increasing the energy consumption\w{, contrary to RSs}. Classical DRL agents on the other hand usually use less energy at the cost of additional comfort violations \w{compared to EAs }in this early phase of learning\del{,} before converging to near-optimal solutions after longer training times~\cite{di2022near}.}

\begin{table}[]
    \caption{Best reward obtained by each agent on the test set over the first $500$ epochs.}
    \vspace{-0.3em}
    \label{tab:final}
    \centering
    \begin{tabular}{lr|lr|lr}
    \hline
        \textbf{Agent} & \textbf{Rew.} & \textbf{Agent} & \textbf{Rew.} & \textbf{Agent} & \textbf{Rew.} \\  \hline
        Classical 1 &	-2.64 &
        Classical 2  &	-2.75	 &
          &		 \\ \hline
        RS\;$0$\;/\;$1$  &	-2.58	 &
        EA\;$0$\;/\;$1$  &	-2.85	 &
        EA\;$0.5$\;/\;$1$ &	-2.83 \\
        RS\;$0$\;/\;$0.5$ &	 -2.44 &
        EA\;$0$\;/\;$0.5$  &	-2.58 &
        EA\;$0.25$\;/\;$0.5$  &	-2.74	 \\
        RS\;$0$\;/\;$0.25$  &	\textbf{\textcolor{blue}{-2.37}}	&
        EA\;$0$\;/\;$0.25$  &	-2.51	&
        EA\;$0.2$\;/\;$0.25$ &	-2.62	 \\
        RS\;$0$\;/\;$0.1$  &	\textcolor{red}{-3.69}  &
        EA\;$0$\;/\;$0.1$ &	\textbf{\textcolor{blue}{-2.46}} &
        EA\;$0.075$\;/\;$0.1$ & \textbf{\textcolor{blue}{-2.46}}\\ \hline
    \end{tabular}
\vspace{-0.5em}
\end{table}

\subsection{Visualization of the impact of \n{prior knowledge}\w{the physical rules}\del{the proposed modifications}}

\begin{figure*}[]
  \begin{center}
    \centerline{\includegraphics[width=0.75\textwidth]{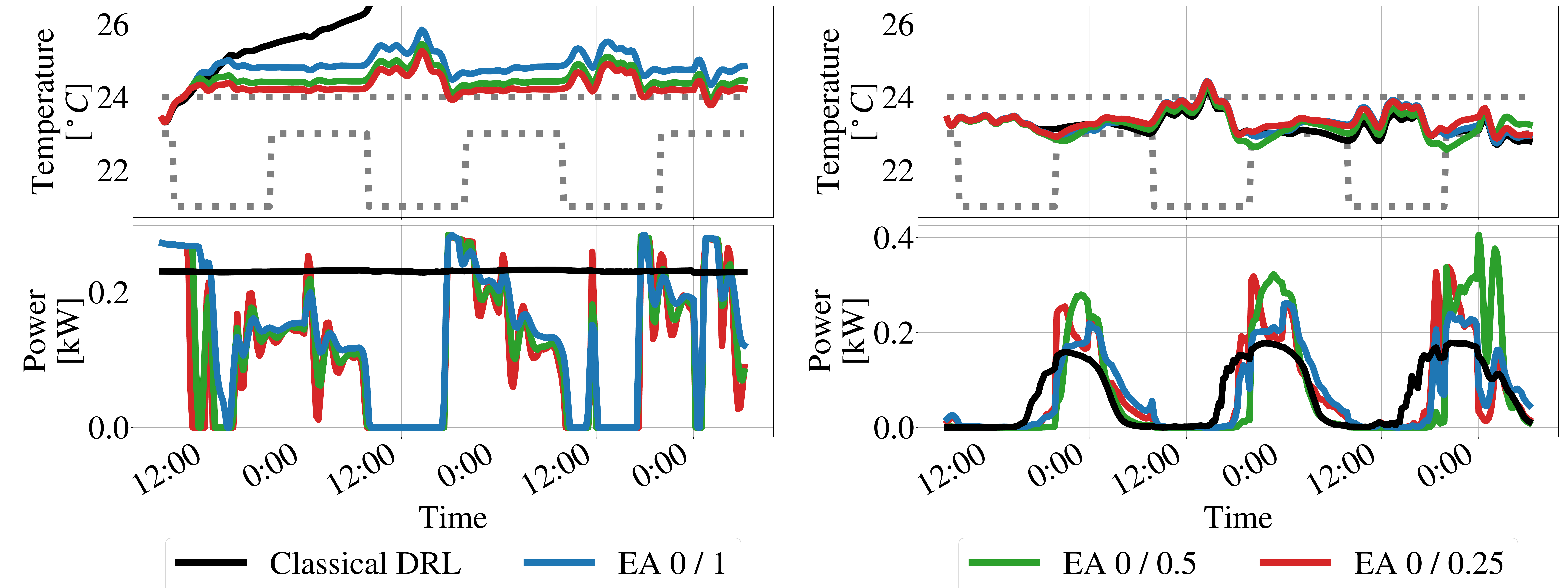}}
  \end{center}
  \vspace{-1.2em}
  \caption{Behavior of a classical \w{DRL }agent and EAs with various $m$ and $n$ parameters (ours) minimizing the \n{heating} power consumption (bottom) while maintaining the temperature in the grey dotted\w{ predefined} bounds (top)\w{ over three days in March}. \textbf{Left:} Performance \textit{before} training\n{, where EAs are saturated once they exceed the bounds by $n$ degrees}. \textbf{Right:} Performance \textit{after} training\n{, showing how all agents converged to similar solutions (Table~\ref{tab:perf}}).} 
  \label{fig:visual}
\vspace{-0.8em}
\end{figure*}

To intuitively understand the effect of action saturation, we\del{ first} visualize its impact on some EAs in Fig.~\ref{fig:visual}, where the behavior of all agents is plotted \textit{before} training on the left, and \textit{after} on the right, for the same three days during the heating season\w{ in March}. Focusing on the left plot, we see the untrained classical DRL agent in black letting the temperature diverge to an uncomfortably high range (out of the bounds of the plot) as it starts exploring the state space using roughly constant heating power. On the other hand, all the EAs are forced to stop heating once they are $n$ degrees out of bounds. Consequently, even before training, such agents will not overheat the room and keep it at acceptable temperatures for the occupants, corresponding to what we expect from good control\new{lers}\del{ policies}. \new{However, note that 
EAs can present control input oscillations}\del{This can however lead to control input oscillations} due to the impact of external disturbances, mainly the solar gains around noon\new{, triggering the saturation mechanism on and off.} \del{which sometimes trigger the overriding mechanism and forces them to suddenly stop heating.}

On the right plot, after training, one can observe that all EAs generally take comparable decisions \new{--- still being sometimes saturated, which ensures compliance with prior expert knowledge ---}  leading to similar temperature patterns\del{, still with some control oscillations stemming from the overriding mechanism}. On the other hand, the classical \del{DRL }agent presents a slightly different behavior, with smoother decision patterns. Interestingly, this agent is the only one heating in the early afternoon, while the EAs wait until the end of the afternoon to heat the room with high power and meet the comfort bound tightening at $8$pm. \del{As presented in Table~\ref{tab:perf}, t}\new{T}his allows the classical agent to use less energy than EAs over these three days but \new{can}\del{it might} incur additional comfort violations \new{(Table~\ref{tab:perf}), as expected from Fig.~\ref{fig:results_app}}. \del{As shown in \del{Appendix}\new{Figure~\ref{fig:results_app}}, these findings generalize to the entire data set: classical DRL agents usually use less energy at the cost of additional comfort violations compared to EAs in the early phase of learning, before converging to near-optimal solutions after longer training times~\cite{di2022near}.  In fact, choosing a smaller $n$, i.e., incorporating more prior knowledge in the EAs, allows them to satisfy the comfort bounds faster without significant additional energy consumption.}

\begin{table}
    \caption{Reward, sum of comfort violations (Vio.), and energy consumption (En.) of each agent over the three days depicted on the right of Fig.~\ref{fig:visual}.}
    \label{tab:perf}
    \vspace{-0.3em}
    \centering
    \begin{tabular}{l|c|c|c|c}
    \hline
        \textbf{Agent} & \textbf{Classical 1} & \multicolumn{3}{c}{\textbf{EA (ours)}} \\ 
        $m$\;/\;$n$ & - & $0$\;/\;$1$ & $0$\;/\;$0.5$ & $0$\;/\;$0.25$ \\ \hline
        Reward & -0.68 & -0.69 & -0.89 & -0.60 \\
        Vio. [Kh] & 1.28 & 1.18 & 2.23 & 0.65 \\
        En. [kWh] & 5.03 & 5.38 & 5.54 & 5.46 \\ \hline
    \end{tabular}
\vspace{-2em}
\end{table}

\subsection{Data efficiency \new{of the proposed gradient modification}}

\begin{figure*}[]
  \begin{center}
    \centerline{\includegraphics[width=0.75\textwidth]{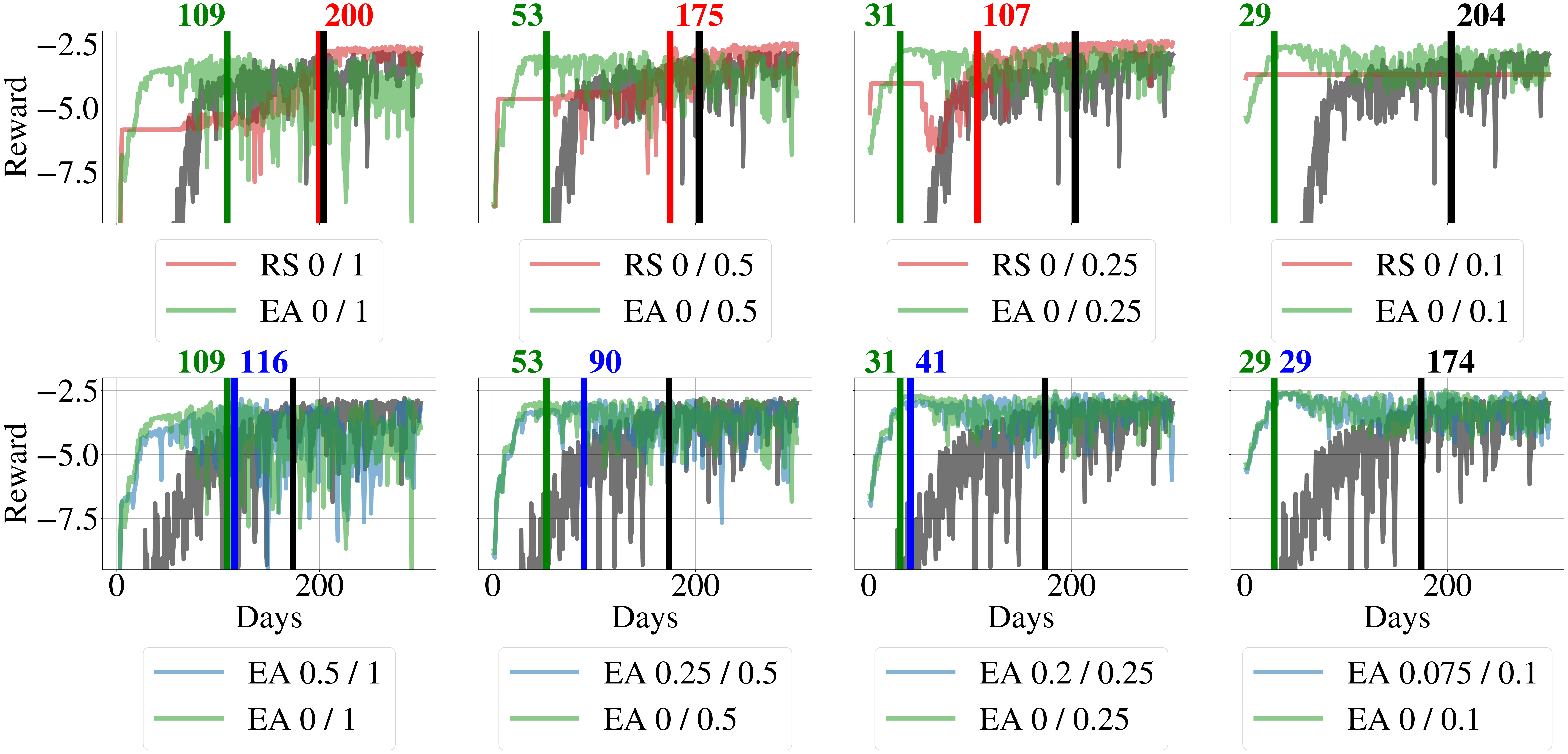}}
  \end{center}
  \vspace{-1.5em}
  \caption{Convergence speed of various EAs with different $m$ and $n$ parameters (ours), compared to agents using RS and two classical agents in black (one in the top plots, one in the bottom ones). The vertical lines and annotations specify the number of days of data required to obtain a reward of $-2.95$ for each agent, which corresponds to the performance of two industrial rule-based baselines.}
    \label{fig:results}
\vspace{-1.5em}
\end{figure*}

A comparison of the convergence speed of various agents over the first $300$ epochs is plotted in Fig.~\ref{fig:results}, where the vertical lines and annotations illustrate the number of days required to attain performance on par with rule-based on-off industrial baselines from \cite{di2022near}. In general, we observe that all the EAs attain returns on par with the baselines significantly earlier than classical DRL agents, in as little as $29$ days instead of roughly $200$, an improvement of almost an order of magnitude. In particular, the smaller $n$ is chosen (from left to right in Fig.~\ref{fig:results}), the faster the convergence of the EAs in green and blue. Intuitively, this makes sense, as tighter constraints introduce more prior knowledge to the EAs, thereby allowing them to find interesting solutions faster, without losing time exploring \new{suboptimal state-action pairs}\del{the state space}. On the other hand, the influence of $m$ is less marked, with $m \neq 0$ (blue) and $m=0$ (green) leading to very similar convergence patterns in the bottom row of plots in Fig.~\ref{fig:results}.

Remarkably, RS does not seem to drastically speed the training up in this case study (red). While RS~$0$\;/\;$0.25$ does converge twice as fast as the classical DRL agents, \del{we can also observe that} RS~\;$0$\;/\;$0.1$ \n{does}\w{did} not converge at all, hinting at the fragility of RS in general. \new{Even when they converge, \n{RSs}\w{they} are still \w{$2$ -- $3$}\n{two to three} times slower than their EA counterparts.} On the other hand, RS seems to lead to more consistent performance than classical agents and EAs after a few hundred epochs, which is confirmed by their good final performance\del{, as detailed in Appendix} \new{in Fig.~\ref{fig:results_app}}. 

Overall, these results support our claim that, as long as the rules provided to the agents are well-defined and correspond to expected behaviors, the proposed modifications can indeed greatly accelerate the convergence of DRL agents. Critically, \rw{however, }this does not significantly impact the quality of the final solution \rw{, as shown in Section}\rn{(Sec.~\ref{sec:perf})}. Interestingly, incorporating more specific expert knowledge in EAs \w{, i.e. using tighter $m$\;/\;$n$ parameters,}\n{--- through smaller $m$ and $n$ ---} further accelerates the\new{ir} convergence\del{ of EAs}. This \rw{again }corresponds to our intuition: \rw{tighter and }better-defined rules help \rw{the }agents more. Remarkably, the modifications proposed in Sec\rn{.}\rw{tion}~\ref{sec:methods} provide the desired speedup for a wide variety of parameters $m$ \n{and} $n$, contrary to RS, hinting at the robustness of the proposed scheme.


\section{Conclusion}

Starting from the postulate that prior expert knowledge often gives us an intuition of how good control policies should behave, we presented a scheme to encode it \w{through simple rules }in \new{actor-critic frameworks}\del{DRL agents} \n{through simple rules} to accelerate \del{the }learning \del{procedure }and decrease the associated computational load. These rules take the form of bounds on the agent's actions\del{ at each time step, which} \new{that} can directly be enforced during \del{offline }training and online operations\del{ without \new{computational} overhead}. \del{Additionally, t}\new{T}o ensure agents learn from their mistakes, we \new{also} modif\new{ied}\del{y} the actor gradients \del{used to update the parameters of the actor }to steer control policies towards expected behaviors, limiting the exploration of \new{known suboptimal state-action pairs}\del{uninteresting states}. \new{Critically, both these operations are computationally inexpensive\rn{, ensuring the gains in sample complexity positively impact the training time of the agents}.}

On a room temperature control case study, this scheme allowed us to accelerate the convergence speed of DRL agents by up to \new{$6$ -- $7\times$}\del{one order of magnitude}. Furthermore, modifying actor gradients proved to be \new{$2$ -- $3$ times} more effective and more robust than the widespread reward shaping method. Remarkably, this was done without suffering from a significant drop in the final performance of the control policies, illustrating how prior knowledge can help \new{alleviate the computational burden of DRL}\del{reduce the computational costs as long as the rules defined by the user match the behavior of optimal policies}. This represents an interesting first step towards \new{efficient}\del{DRL} agents that can be deployed and learned from scratch on physical systems, potentially bypassing the need for complex simulators.  In future work, it would\del{ hence} be interesting to investigate annealing strategies on $\lambda$ \new{or leverage primal-dual optimization tools} to \new{adaptively}\del{slowly} \del{alleviate}\new{tune} the \new{influence of the} additional penalty in the actor gradient and let agents learn more expressive policies \new{after}\del{once} the initial exploration phase\del{ has been carried on}. 


{
\bibliographystyle{IEEEtran}
\bibliography{biblio}
}



\del{The best reward obtained by all the trained agents over the first $500$ epochs can be found in Table~\ref{tab:final}, and the corresponding trade-off between energy consumption and comfort violations is plotted in Fig.~\ref{fig:results_app}. These snapshot results after $500$ epochs also illustrate how tighter constraints, i.e., higher levels of prior knowledge, allow EAs and RSs to converge to better solutions in this limited training regime.}

\end{document}